\documentclass{article}
\usepackage{arxiv}
\usepackage[utf8]{inputenc} %
\usepackage[T1]{fontenc}    %
\usepackage{hyperref}       %
\usepackage{url}            %
\usepackage{booktabs}       %
\usepackage{amsfonts}       %
\usepackage{nicefrac}       %
\usepackage{microtype}      %
\usepackage{amsmath}
\usepackage[dvipsnames]{xcolor}
\usepackage{lipsum}		%
\usepackage{graphicx}
\usepackage{multirow}
\usepackage[square,numbers]{natbib}
\usepackage{doi}
\usepackage{siunitx}
\usepackage{enumitem}
\usepackage{acronym}

\title{The Role of Temporal Hierarchy in Spiking Neural Networks}

\usepackage{authblk}

\newbox{\orcid}\sbox{\orcid}{\includegraphics[scale=0.06]{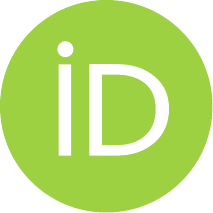}} 
\author[*,1]{%
	\href{https://orcid.org/0000-0002-5279-6309}{\usebox{\orcid}\hspace{1mm}Filippo Moro}%
}
\author[1]{%
	\href{https://orcid.org/0000-0002-1218-4009}{\usebox{\orcid}\hspace{1mm}Pau Vilimelis Aceituno}%
 }
\author[1]{%
	\href{https://orcid.org/0000-0001-5275-9199}{\usebox{\orcid}\hspace{1mm}Laura Kriener}%
 }
\author[1]{%
	\href{https://orcid.org/0000-0001-5400-067X}{\usebox{\orcid}\hspace{1mm}Melika Payvand}%
} 
\vspace{1cm}

\affil[1]{Institute of Neuroinformatics, UZH and ETH Zurich, Zurich, CH}
\affil[*]{Corresponding author: filippo.moro@uzh.ch}

\date{}

\hypersetup{
pdftitle={The role of temporal hierarchy in Spiking Neural Networks},
pdfauthor={F. Moro et al.},
pdfkeywords={Hierarchy, SNN},
}

\begin{document}
\acrodef{AI}[AI]{Artificial Intelligence}
\acrodef{ADC}[ADC]{Analog to Digital Converter}
\acrodef{ADEXP}[AdExp-I\&F]{Adaptive-Exponential Integrate and Fire}
\acrodef{AER}[AER]{Address-Event Representation}
\acrodef{AEX}[AEX]{AER EXtension board}
\acrodef{AE}[AE]{Address-Event}
\acrodef{AFM}[AFM]{Atomic Force Microscope}
\acrodef{AGC}[AGC]{Automatic Gain Control}
\acrodef{AMDA}[AMDA]{AER Motherboard with D/A converters}
\acrodef{ANN}[ANN]{Artificial Neural Network}
\acrodef{API}[API]{Application Programming Interface}
\acrodef{ARM}[ARM]{Advanced RISC Machine}
\acrodef{ASIC}[ASIC]{Application Specific Integrated Circuit}
\acrodef{AdExp}[AdExp-IF]{Adaptive Exponential Integrate-and-Fire}
\acrodef{BCM}[BMC]{Bienenstock-Cooper-Munro}
\acrodef{BD}[BD]{Bundled Data}
\acrodef{BEOL}[BEOL]{Back-end of Line}
\acrodef{BG}[BG]{Bias Generator}
\acrodef{BMI}[BMI]{Brain-Machince Interface}
\acrodef{BPTT}[BPTT]{Backpropagation Through Time}
\acrodef{BTB}[BTB]{band-to-band tunnelling}
\acrodef{CAD}[CAD]{Computer Aided Design}
\acrodef{CAM}[CAM]{Content Addressable Memory}
\acrodef{CAVIAR}[CAVIAR]{Convolution AER Vision Architecture for Real-Time}
\acrodef{CA}[CA]{Cortical Automaton}
\acrodef{CCN}[CCN]{Cooperative and Competitive Network}
\acrodef{CDR}[CDR]{Clock-Data Recovery}
\acrodef{CFC}[CFC]{Current to Frequency Converter}
\acrodef{CHP}[CHP]{Communicating Hardware Processes}
\acrodef{CMIM}[CMIM]{Metal-insulator-metal Capacitor}
\acrodef{CML}[CML]{Current Mode Logic}
\acrodef{CMOL}[CMOL]{Hybrid CMOS nanoelectronic circuits}
\acrodef{CMOS}[CMOS]{Complementary Metal-Oxide-Semiconductor}
\acrodef{CNN}[CCN]{Convolutional Neural Network}
\acrodef{COTS}[COTS]{Commercial Off-The-Shelf}
\acrodef{CPG}[CPG]{Central Pattern Generator}
\acrodef{CPLD}[CPLD]{Complex Programmable Logic Device}
\acrodef{CPU}[CPU]{Central Processing Unit}
\acrodef{CSM}[CSM]{Cortical State Machine}
\acrodef{CSP}[CSP]{Constraint Satisfaction Problem}
\acrodef{CV}[CV]{Coefficient of Variation}
\acrodef{DAC}[DAC]{Digital to Analog Converter}
\acrodef{DAS}[DAS]{Dynamic Auditory Sensor}
\acrodef{DAVIS}[DAVIS]{Dynamic and Active Pixel Vision Sensor}
\acrodef{DBN}[DBN]{Deep Belief Network}
\acrodef{DFA}[DFA]{Deterministic Finite Automaton}
\acrodef{DIBL}[DIBL]{drain-induced-barrier-lowering}
\acrodef{DI}[DI]{delay insensitive}
\acrodef{DMA}[DMA]{Direct Memory Access}
\acrodef{DNF}[DNF]{Dynamic Neural Field}
\acrodef{DNN}[DNN]{Deep Neural Network}
\acrodef{DOF}[DOF]{Degrees of Freedom}
\acrodef{DPE}[DPE]{Dynamic Parameter Estimation}
\acrodef{DPI}[DPI]{Differential Pair Integrator}
\acrodef{DRRZ}[DR-RZ]{Dual-Rail Return-to-Zero}
\acrodef{DRAM}[DRAM]{Dynamic Random Access Memory}
\acrodef{DR}[DR]{Dual Rail}
\acrodef{DSP}[DSP]{Digital Signal Processor}
\acrodef{DVS}[DVS]{Dynamic Vision Sensor}
\acrodef{DYNAP}[DYNAP]{Dynamic Neuromorphic Asynchronous Processor}
\acrodef{EBL}[EBL]{Electron Beam Lithography}
\acrodef{EDVAC}[EDVAC]{Electronic Discrete Variable Automatic Computer}
\acrodef{EEG}[EEG]{Electroencephalography}
\acrodef{ECG}[ECG]{Electrocardiography}
\acrodef{EMG}[EMG]{Electromyography}
\acrodef{EIN}[EIN]{Excitatory-Inhibitory Network}
\acrodef{EM}[EM]{Expectation Maximization}
\acrodef{EPSC}[EPSC]{Excitatory Post-Synaptic Current}
\acrodef{EPSP}[EPSP]{Excitatory Post-Synaptic Potential}
\acrodef{ESN}[ESN]{Echo state Network }
\acrodef{EZ}[EZ]{Epileptogenic Zone}
\acrodef{FDSOI}[FDSOI]{Fully-Depleted Silicon on Insulator}
\acrodef{FET}[FET]{Field-Effect Transistor}
\acrodef{FFT}[FFT]{Fast Fourier Transform}
\acrodef{FI}[F-I]{Frequency-Current}
\acrodef{FPGA}[FPGA]{Field Programmable Gate Array}
\acrodef{FR}[FR]{Fast Ripple}
\acrodef{FSA}[FSA]{Finite State Automaton}
\acrodef{FSM}[FSM]{Finite State Machine}
\acrodef{GIDL}[GIDL]{gate-induced-drain-leakage}
\acrodef{GOPS}[GOPS]{Giga-Operations per Second}
\acrodef{GPU}[GPU]{Graphical Processing Unit}
\acrodef{GUI}[GUI]{Graphical User Interface}
\acrodef{HAL}[HAL]{Hardware Abstraction Layer}
\acrodef{HFO}[HFO]{High Frequency Oscillation}
\acrodef{HH}[H\&H]{Hodgkin \& Huxley}
\acrodef{HMM}[HMM]{Hidden Markov Model}
\acrodef{HCS}[HCS]{High-Conductive State}
\acrodef{HRS}[HRS]{High-Resistive State}
\acrodef{HR}[HR]{Human Readable}
\acrodef{HSE}[HSE]{Handshaking Expansion}
\acrodef{HW}[HW]{Hardware}
\acrodef{ICT}[ICT]{Information and Communication Technology}
\acrodef{IC}[IC]{Integrated Circuit}
\acrodef{IEEG}[iEEG]{intracranial electroencephalography}
\acrodef{IF2DWTA}[IF2DWTA]{Integrate \& Fire 2--Dimensional WTA}
\acrodef{IFSLWTA}[IFSLWTA]{Integrate \& Fire Stop Learning WTA}
\acrodef{IF}[I\&F]{Integrate-and-Fire}
\acrodef{IMU}[IMU]{Inertial Measurement Unit}
\acrodef{INCF}[INCF]{International Neuroinformatics Coordinating Facility}
\acrodef{INI}[INI]{Institute of Neuroinformatics}
\acrodef{IO}[I/O]{Input/Output}
\acrodef{IPSC}[IPSC]{Inhibitory Post-Synaptic Current}
\acrodef{IPSP}[IPSP]{Inhibitory Post-Synaptic Potential}
\acrodef{IP}[IP]{Intellectual Property}
\acrodef{ISI}[ISI]{Inter-Spike Interval}
\acrodef{IoT}[IoT]{Internet of Things}
\acrodef{JFLAP}[JFLAP]{Java - Formal Languages and Automata Package}
\acrodef{LEDR}[LEDR]{Level-Encoded Dual-Rail}
\acrodef{LFP}[LFP]{Local Field Potential}
\acrodef{LIF}[LIF]{Leaky Integrate and Fire}
\acrodef{LLC}[LLC]{Low Leakage Cell}
\acrodef{LNA}[LNA]{Low-Noise Amplifier}
\acrodef{LPF}[LPF]{Low Pass Filter}
\acrodef{LCS}[LCS]{Low-Conductive State}
\acrodef{LRS}[LRS]{Low-Resistive State}
\acrodef{LSM}[LSM]{Liquid State Machine}
\acrodef{LTD}[LTD]{Long Term Depression}
\acrodef{LTI}[LTI]{Linear Time-Invariant}
\acrodef{LTP}[LTP]{Long Term Potentiation}
\acrodef{LTU}[LTU]{Linear Threshold Unit}
\acrodef{LUT}[LUT]{Look-Up Table}
\acrodef{LVDS}[LVDS]{Low Voltage Differential Signaling}
\acrodef{MCMC}[MCMC]{Markov-Chain Monte Carlo}
\acrodef{MEMS}[MEMS]{Micro Electro Mechanical System}
\acrodef{MFR}[MFR]{Mean Firing Rate}
\acrodef{MIM}[MIM]{Metal Insulator Metal}
\acrodef{MLP}[MLP]{Multilayer Perceptron}
\acrodef{MOSCAP}[MOSCAP]{Metal Oxide Semiconductor Capacitor}
\acrodef{MOSFET}[MOSFET]{Metal Oxide Semiconductor Field-Effect Transistor}
\acrodef{MOS}[MOS]{Metal Oxide Semiconductor}
\acrodef{MRI}[MRI]{Magnetic Resonance Imaging}
\acrodef{NDFSM}[NDFSM]{Non-deterministic Finite State Machine} 
\acrodef{ND}[ND]{Noise-Driven}
\acrodef{NEF}[NEF]{Neural Engineering Framework}
\acrodef{NHML}[NHML]{Neuromorphic Hardware Mark-up Language}
\acrodef{NIL}[NIL]{Nano-Imprint Lithography}
\acrodef{NMDA}[NMDA]{N-Methyl-D-Aspartate}
\acrodef{NME}[NE]{Neuromorphic Engineering}
\acrodef{NN}[NN]{Neural Network}
\acrodef{NRZ}[NRZ]{Non-Return-to-Zero}
\acrodef{NSM}[NSM]{Neural State Machine}
\acrodef{OR}[OR]{Operating Room}
\acrodef{OTA}[OTA]{Operational Transconductance Amplifier}
\acrodef{PCB}[PCB]{Printed Circuit Board}
\acrodef{PCHB}[PCHB]{Pre-Charge Half-Buffer}
\acrodef{PCM}[PCM]{Phase Change Memory}
\acrodef{PE}[PE]{Processing Element}
\acrodef{PFA}[PFA]{Probabilistic Finite Automaton}
\acrodef{PFC}[PFC]{prefrontal cortex}
\acrodef{PFM}[PFM]{Pulse Frequency Modulation}
\acrodef{PR}[PR]{Production Rule}
\acrodef{PSC}[PSC]{Post-Synaptic Current}
\acrodef{PSP}[PSP]{Post-Synaptic Potential}
\acrodef{PSTH}[PSTH]{Peri-Stimulus Time Histogram}
\acrodef{QDI}[QDI]{Quasi Delay Insensitive}
\acrodef{RAM}[RAM]{Random Access Memory}
\acrodef{RDF}[RDF]{random dopant fluctuation}
\acrodef{RELU}[ReLu]{Rectified Linear Unit}
\acrodef{RLS}[RLS]{Recursive Least-Squares}
\acrodef{RMSE}[RMSE]{Root Mean Squared-Error}
\acrodef{RMS}[RMS]{Root Mean Squared}
\acrodef{RNN}[RNN]{Recurrent Neural Networks}
\acrodef{RSNN}[RSNN]{Recurrent Spiking Neural Network}
\acrodef{ROLLS}[ROLLS]{Reconfigurable On-Line Learning Spiking}
\acrodef{RRAM}[RRAM]{Resistive Random Access Memory}
\acrodef{R}[R]{Ripples}
\acrodef{SAC}[SAC]{Selective Attention Chip}
\acrodef{SAT}[SAT]{Boolean Satisfiability Problem}
\acrodef{SCX}[SCX]{Silicon CorteX}
\acrodef{SD}[SD]{Signal-Driven}
\acrodef{SEM}[SEM]{Spike-based Expectation Maximization}
\acrodef{SLAM}[SLAM]{Simultaneous Localization and Mapping}
\acrodef{SNN}[SNN]{Spiking Neural Network}
\acrodef{SNR}[SNR]{Signal to Noise Ratio}
\acrodef{SOC}[SOC]{System-On-Chip}
\acrodef{SOI}[SOI]{Silicon on Insulator}
\acrodef{SP}[SP]{Separation Property}
\acrodef{SHD}[SHD]{Spiking Heidelberg Digit}
\acrodef{SRAM}[SRAM]{Static Random Access Memory}
\acrodef{SRNN}[SRNN]{Spiking Recurrent Neural Network}
\acrodef{STDP}[STDP]{Spike-Timing Dependent Plasticity}
\acrodef{STD}[STD]{Short-Term Depression}
\acrodef{STP}[STP]{Short-Term Plasticity}
\acrodef{STT-MRAM}[STT-MRAM]{Spin-Transfer Torque Magnetic Random Access Memory}
\acrodef{STT}[STT]{Spin-Transfer Torque}
\acrodef{SW}[SW]{Software}
\acrodef{TCAM}[TCAM]{Ternary Content-Addressable Memory}
\acrodef{TFT}[TFT]{Thin Film Transistor}
\acrodef{TPU}[TPU]{Tensor Processing Unit}
\acrodef{USB}[USB]{Universal Serial Bus}
\acrodef{VHDL}[VHDL]{VHSIC Hardware Description Language}
\acrodef{VLSI}[VLSI]{Very Large Scale Integration}
\acrodef{VOR}[VOR]{Vestibulo-Ocular Reflex}
\acrodef{WCST}[WCST]{Wisconsin Card Sorting Test}
\acrodef{WTA}[WTA]{Winner-Take-All}
\acrodef{XML}[XML]{eXtensible Mark-up Language}
\acrodef{CTXCTL}[CTXCTL]{CortexControl}
\acrodef{divmod3}[DIVMOD3]{divisibility of a number by three}
\acrodef{hWTA}[hWTA]{hard Winner-Take-All}
\acrodef{sWTA}[sWTA]{soft Winner-Take-All}
\acrodef{APMOM}[APMOM]{Alternate Polarity Metal On Metal}
\acrodef{SRNN}[SRNN]{Spiking Recurrent Neural Networks}
\acrodef{fMRI}[fMRI]{functional Magnetic Resonance Imaging}
\acrodef{RL}[RL]{Reinforcement Learning}
\acrodef{ES}[ES]{Evolutionary Strategies}

\maketitle

\begin{abstract}

\acp{SNN} have the potential for rich spatio-temporal signal processing thanks to exploiting both spatial and temporal parameters. The temporal dynamics such as time constants of the synapses and neurons and delays have been recently shown to have computational benefits that help reduce the overall number of parameters required in the network and increase the accuracy of the \acp{SNN} in solving temporal tasks. Optimizing such temporal parameters, for example, through gradient descent, gives rise to a temporal architecture for different problems. As has been shown in machine learning, to reduce the cost of optimization, architectural biases can be applied, in this case in the temporal domain. Such inductive biases in temporal parameters have been found in neuroscience studies, highlighting a hierarchy of temporal structure and input representation in different layers of the cortex. 
Motivated by this, we propose to impose a hierarchy of temporal representation in the hidden layers of \acp{SNN}, highlighting that such an inductive bias improves their performance. We demonstrate the positive effects of temporal hierarchy in the time constants of feed-forward \acp{SNN} applied to temporal tasks (Multi-Time-Scale XOR and Keyword Spotting, with a benefit of up to 4.1\% in classification accuracy). 
Moreover, we show that such architectural biases, i.e. hierarchy of time constants, naturally emerge when optimizing the time constants through gradient descent, initialized as homogeneous values.
We further pursue this proposal in temporal convolutional \acp{SNN}, by introducing the hierarchical bias in the size and dilation of temporal kernels, giving rise to competitive results in popular temporal spike-based datasets.
\end{abstract}


\section{Introduction}

\acp{SNN} utilize time as an additional dimension for computation and have thus been shown promising for temporal data processing \cite{eshraghian2023training}. As their computational unit, \acp{SNN} incorporate spiking neurons that integrate information over time in their cell body and generate an electric pulse, i.e., a spike, when this value passes a threshold.
Arguably, the spikes convey this information through the timing of their generation, or their frequency~\cite{foffani2009spike}. 
Different models of spiking neurons have been developed, from simplified \ac{LIF} models to more biologically realistic ones \cite{gerstner2002spiking}. 
To perform computation, \acp{SNN} exploit the dynamics of such spiking neurons for generating short- and long-term memory \cite{bellec2018long}.

\begin{figure}[t]
    \centering
    \includegraphics[width=0.8\linewidth]{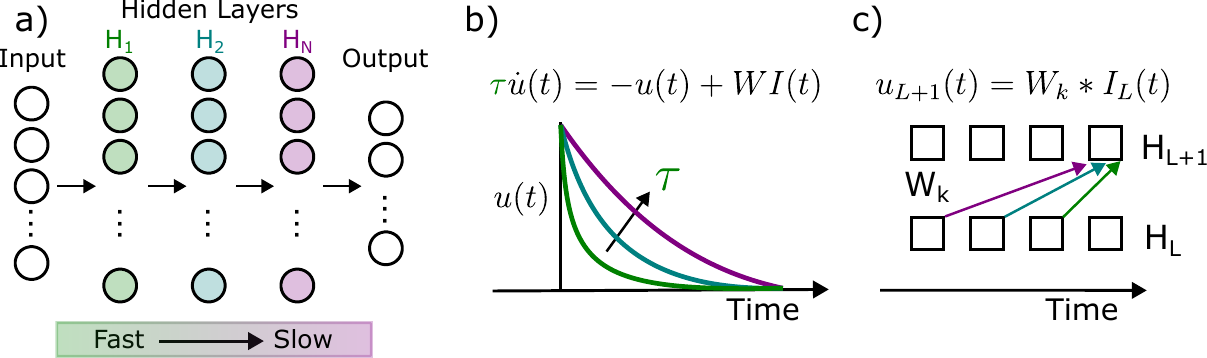}
    \caption{Temporal hierarchy in Spiking Neural Networks (SNNs). a) Multi-Layer SNN featuring $\mathcal{N}$ hidden layers, each colored differently, highlighting the difference in temporal processing speed of the neurons in the hidden layers, from Fast in the first hidden layers to Slow in the deeper layers. b) Time constant $\mathbf{\tau}$ determining the speed of neuronal dynamics in Leaky-Integrate-and-Fire (LIF) spiking neurons. The differential equation describes the dynamics for the membrane voltage $u(t)$, which evolves with a time constant $\mathbf{\tau}$. c) Temporal processing expressed in temporal convolutions by varying the weights' kernel $W_k$, affecting the state of neurons $u(t)$ in time.}
    \label{fig:Figure1}
\end{figure}

It has been shown that optimizing these dynamics in \acp{SNN} increases the accuracy and/or reduces the number of parameters of the \acp{SNN} in temporal tasks. 
These optimized dynamics include the leaky dynamics of the neurons (i.e. their time constant) \cite{yin2023accurate}, delay parameters in the synapses~\cite{d2024denram, goltz2024delgrad} or temporal causal convolutions~\cite{hammouamri2023learning}, adaptation of the threshold voltage of the neurons~\cite{bellec2018long, bittar2022surrogate, yin2020effective}, and recurrent connections resulting in short-term memory~\cite{graves2013generating, khajehabdollahi2023emergent}. 

While these computational principles are commonly featured in \acp{SNN}, reaching their full potential remains an open research problem. 
In this work, we take a step in combining understandings from the machine learning and computational neuroscience literature to further exploit the temporal dynamics of SNN for better performance.

On the one hand, the success of Deep Neural Networks (DNNs) has shown the computational benefits of hierarchical representation of inputs in successive deep layers~\cite{srivastava2015training, telgarsky2016benefits}, and recently, such hierarchical input representation has also been applied to temporal data~\cite{weidel2021wavesense}.

On the other hand, recent neuroscience findings highlight the beneficial role of heterogeneity introduced to the dynamics of \acp{SNN}. For example, it has been shown that exploiting a variety of time constants in a population of neurons improves the performance of \acp{SNN}~\cite{perez2021neural, zheng2024temporal}. 
Moreover, the concept of the hierarchy in structure \cite{harris2019hierarchical, meunier2009hierarchical}, input and action representation \cite{wacongne2011evidence, grafton2007evidence} in the brain is well studied in neuroscience. Authors in \cite{honey2012slow} show that the hierarchical organization in the cortex is not only used for spatial operations but also for temporal operations. They also prove that early sensory regions tend to be affected by more recent input features, while the activity of higher-order cortical regions correlates with longer windows of prior stimulus context. 
The timescales of fluctuation in spiking activity have been observed to increase across the hierarchy of cortical areas in primates \cite{murray2014hierarchy}. This evidence hints that temporal processing is organized in a hierarchy in the brain, and this might be crucial to developing complex behavior.

Combining these lessons from DNNs and computational neuroscience, i.e. benefits of hierarchical representation and heterogeneity, we analyze the implications of imposing a hierarchy of temporal processing speed in \acp{SNN}.
We propose to leverage two important temporal processing elements of \acp{SNN}, neural dynamics in the form of the time constant and delays or temporal causal convolutions. With these elements, we construct a hierarchy of temporal representation in multi-layer \acp{SNN} (Figure \ref{fig:Figure1}a) that brings a performance benefit in common spike-based benchmark tasks. Also, we show that performance decreases if the hierarchy of the temporal representation is reversed. In general, we advocate for a simple setting in initializing \acp{SNN}, whereby temporal processing elements are ordered from fast to slow through the hidden layers of the SNN.

Different sets of experiments are performed to prove the hypothesis on the benefit of temporal hierarchy, testing \acp{SNN} on common spike-based benchmark tasks, such as the Multi-Time-Scale XOR \cite{zheng2024temporal}, the \ac{SHD} \cite{cramer2020heidelberg} and the spike-based version of the Google Speech Command dataset (SSC) \cite{warden2018speech, cramer2020heidelberg}. 
First, the time constants of neurons are initialized in a crescendo through the hidden layers of \acp{SNN}. This proves to be a useful inductive bias in forming a complex representation of the input and improving its computational power of the \acp{SNN}, with a performance increase of up to 4.1\% in the Spiking Heidelberg Digits dataset. Additionally, we find that such temporal hierarchy naturally emerges through optimization, as a result of training the time constants of the neurons of an SNN with backpropagation.
We also demonstrate that the introduction of temporal hierarchy is useful in \acp{SNN} based on temporal causal convolutions, where the kernel and dilation of the convolutions are arranged hierarchically in the network, reaching very competitive classification accuracy in keyword-spotting datasets, up to 94.1\% on the SHD and 79.2\% on the SSC.\\
Overall, our results shed light on the role of the temporal hierarchy as a positive inductive bias for \acp{SNN} and provide a simple heuristic to improve the performance of \acp{SNN} in temporal tasks.

\section{Results} \label{sec:headings}

To verify the hypothesis that a hierarchy of temporal processing is beneficial to \acp{SNN}, we set up a series of experiments involving common benchmark tasks. In particular, the benefit is expected when the SNN processes data with a temporal structure. To prove this, we choose three relevant temporal tasks: the multi-time-scale (MTS) XOR \cite{zheng2024temporal}, the Spiking Heidelberg Digits (Keyword Spotting) \cite{cramer2020heidelberg}, and the Spiking Speech Command (Keyword Spotting) \cite{cramer2020heidelberg, warden2018speech}. Additionally, we investigate the role of temporal hierarchy with static data such as images (see Supplementary Figure \ref{fig:FigureS1}a). As temporal features do not play a role in such a task, temporal hierarchy is expected not to provide added benefits. However, it is possible to revert static images to sequences, by feeding \acp{SNN} single pixels at a time. In this case, temporal features emerge and temporal hierarchy should boost \acp{SNN}' performance.

Temporal processing can be expressed in multiple forms in \acp{SNN}. Neuronal dynamics is the most characteristic one in \acp{SNN}. Lately, delays (or causal temporal convolutions) emerged as a promising temporal processing element, allowing \acp{SNN} to improve performance considerably~\cite{goltz2024delgrad,hammouamri2023learning,d2024denram}. 
As such, we express the concept of temporal hierarchy in two main ways: first, as the difference of the time constant in the hidden layers of \acp{SNN} (Figure \ref{fig:Figure1}b); second, as the varying kernel size and dilation of causal temporal convolutions in \acp{SNN} (Figure \ref{fig:Figure1}c). In both cases, we aim to show that arranging the temporal computational features from fast to slow, through the hidden layers of \acp{SNN}, is beneficial for performance. The experiments are organized in three sets, each providing intuition on the role of temporal hierarchy in \acp{SNN}:
\begin{enumerate}[label=(\roman*)]
    \item Hierarchy of time constant in feed-forward \acp{SNN}
    \item Discovering hierarchy via the optimization of time constants
    \item Temporal hierarchy in convolutional \acp{SNN}
\end{enumerate}

Along with the experiments, we compare the benefits of temporal hierarchy with other inductive biases in \acp{SNN}, confronting our results with the state of the art.

\begin{figure}[t!]
    \centering
    \includegraphics[width=0.95\linewidth]{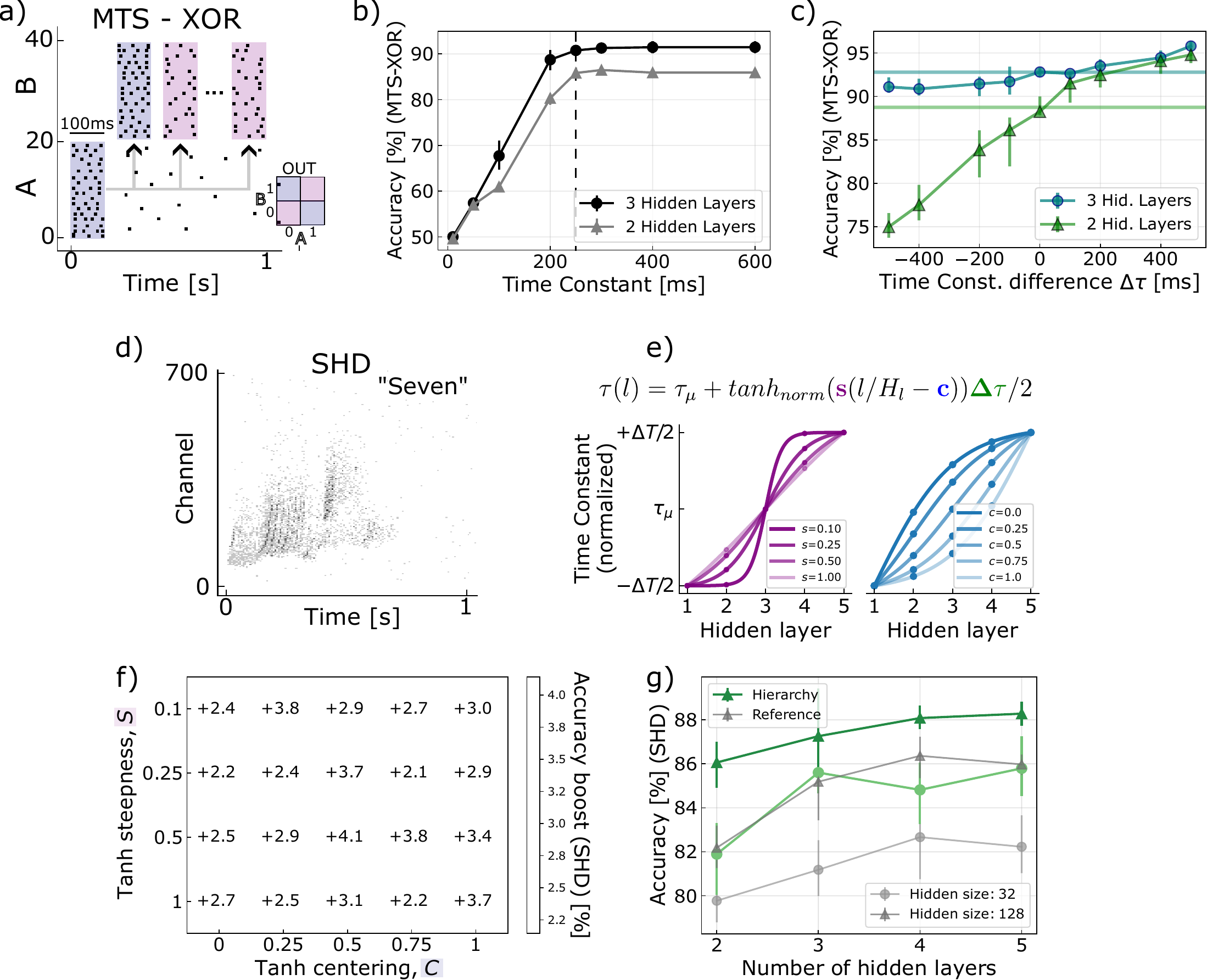}
    \caption{Temporal hierarchy through the time constants of \acp{SNN}. a) Sketch of the Multi-Time-Scale XOR (MTS-XOR) task. b) Classification accuracy on MST-XOR as a function of the homogeneous time constant through the hidden layers. c) Classification accuracy when temporal hierarchy is introduced in multi-layer \acp{SNN} with varying amplitude ($\color{ForestGreen}\Delta \tau$). Positive $\color{ForestGreen}\Delta \tau$ improves the classification accuracy. d) Sample from the Spiking Heidelberg Digits (SHD) dataset. e) The time constant hierarchy is parametrized according to a tanh function, with steepness ($\color{Purple}s$) and centering ($\color{blue}c$) parameters. f) The effect of such parameters is tested on the SHD task, with ${\color{ForestGreen}\Delta \tau}=\SI[retain-explicit-plus]{+150}{\milli\second}$. In all cases, hierarchy has a positive effect compared to the reference performance (${\color{ForestGreen}\Delta \tau }=\SI{0}{\milli\second}$). g) Temporal hierarchy is applied to networks of varying size, showing that an SNN with 32 hidden neurons per layer and temporal hierarchy performs as well as a bigger SNN with 128 neurons per layer without temporal hierarchy. Error bars show the quartiles over 10 trials in all plots.}
    \label{fig:Figure2}
\end{figure}

\subsection{Time constant hierarchy in Feed-Forward SNN} \label{hierarchy_tau}
Neuronal dynamics is the most characteristic temporal processing feature in \acp{SNN}; the spectrum of possible neuron models ranges from more bio-realist multi-compartment implementations~\cite{stefanou2016creating}, to simplified point-like structures e.g., used in McCulloch-Pitts. 
The \ac{LIF} model has been widely adopted in the neuromorphic community for its trade-off between expressivity of neuronal dynamics, computational complexity, and ease of hardware implementation. 
The \ac{LIF} model features a state variable - often called membrane voltage $u(t)$ - which evolves in time according to a first-order differential equation (Figure~\ref{fig:Figure1}b). The temporal processing of \ac{LIF} neurons is expressed by the time constant $\tau$ of the membrane voltage, which enables them to detect the coincidence between two input spikes, aka Coincidence Detection \cite{masquelier2018stdp, moro2022neuromorphic}, and to filter input spike patterns based on their precise timing \cite{goltz2021fast}. 
In this analysis, we examine the effect of implementing a temporal hierarchy via time constants in feed-forward \acp{SNN}. The time constants of \ac{LIF} neurons are set to be fast in the initial layers and slow in the deeper layers. We use a feed-forward network to isolate the impact of the time-constant hierarchy. 
While recurrent networks can control neuron ``speed'' and memory by training recurrent weights, we demonstrate that the resulting working memory can be achieved through the neuron's inherent dynamics alone. 
Here, the time constants are not trained after initialization; this aspect will be addressed in section \ref{optimize_tau}.

\paragraph*{MTS-XOR task}
We train feed-forward \acp{SNN} with 2 and 3 hidden layers on the MTS-XOR dataset \cite{zheng2024temporal}, each with 10 neurons per hidden layer.
The MTS-XOR (Fig.~\ref{fig:Figure2}a) is a synthetic dataset where a single cue is provided at $t=\SI{0}{\second}$ from one of two channels (Channel A) - composed of a population of neurons - with multiple other cues successively provided from the other channel (Channel B) in time. Cues are binary activations of \SI{20}{\milli\second} each expressed as Poisson firing rates, and the task consists of performing the XOR operation each time a cue is generated from Channel B, based on the original cue in Channel A, which is not repeated. The total duration of each datapoint is \SI{1}{\second} and the time-step is set to \SI{10}{\milli \second}. Each cue is presented for \SI{100}{\milli \second} with a Poisson frequency of either \SI{10}{\hertz} (activation 1) or \SI{10}{\kilo\hertz} (activation 0). A break of \SI{50}{\milli \second} is interposed between cues. The task can be successively solved if the network holds the information of the first cue while being able to react quickly to the new cues and perform the XOR operation.
In our preliminary analysis, we set the time constants in a 2- or 3-hidden layer \ac{SNN} to a fixed value, $\tau_{\mu}$, which we varied to identify the optimal value for the MTS-XOR task. Figure~\ref{fig:Figure2}b shows that a time constant of \SI{300}{\milli\second} is sufficient to achieve maximum classification accuracy.
Fixing the average time constant to $\tau_{\mu}= \SI{300}{\milli \second}$, we then modify the time constant of hidden layers $\tau(l)$ to create a hierarchy of time constant proportional to $\color{ForestGreen}\Delta \tau$.
\begin{equation*}
    \tau(l) = \tau_{\mu} + (l - H_l /2) {\color{ForestGreen} \Delta \tau} /2
\end{equation*}
We then train \acp{SNN} with different values of $\color{ForestGreen}\Delta \tau$ in Fig.\ref{fig:Figure2}c. Following the hypothesis on temporal hierarchy, we show that classification accuracy positively correlates with the hierarchy amplitude $\color{ForestGreen}\Delta \tau$, i.e. the time constant difference across the hidden layers of the feed-forward \ac{SNN}. 
In an \ac{SNN} with 2 hidden layers, having a fast first layer and a slow last layer is beneficial. Adding a third hidden layer generally improves performance and makes the impact of temporal hierarchy less prominent, yet the correlation of performance with the hierarchy amplitude persists. 
At the same time, reversing the order of the hierarchy reduces the computational power of the \acp{SNN}. 
We hypothesize that this could be due to the functional role of different layers in a multi-layer \ac{SNN}, with the first hidden layer acting as a feature extractor, while the deeper hidden layers integrate information over longer time spans. 
To verify this, we performed additional experiments in the Supplementary Information \ref{Supplementary_FF_SNN}. 
As the MTS-XOR dataset includes background noise, this hypothesis indicates that having a first fast hidden layer would help filter out such noise, while the information of the first cue is ``stored'' in the slow dynamics of the second layer. 
If the first hidden layer was slower, the noise would be integrated and the memory of the first cue would be lost. 
This is confirmed in the Supplementary Figure \ref{fig:FigureS1}a, where the classification accuracy in the MTS-XOR task is analyzed as a function of the background noise rate. Positive temporal hierarchy (fast to slow dynamics) makes the \ac{SNN} more resilient to noise.

\paragraph*{Keyword Spotting}
In the second set of experiments temporal hierarchy is tested on the \ac{SHD} task (Fig.~\ref{fig:Figure2}d), a common benchmark in the spike-based processing domain, consisting of audio recordings of spoken digits in two languages (English and German) converted to trains of spikes. Each spoken digit is recorded for about \SI{1}{\second} and at \SI{16}{\kilo\hertz} and converted to the spike domain with a bio-plausible cochlear model \cite{cramer2020heidelberg}.

Given that \ac{SHD} is a more complex task than MTS-XOR, we utilize a deeper \ac{SNN} with up to 5 hidden layers, with 32 neurons per hidden layer. 
With more layers available, the time constant for each layer is determined with a hyperbolic tangent $\tanh$ function, to explore the optimal ``shape'' of temporal hierarchy.
\begin{equation*}
    \tau(l) = \tau_{\mu} + \tanh_\mathrm{norm}\left( {\color{Purple} s}(l/H_l-{\color{blue} c}\right) ) {\color{ForestGreen} \Delta \tau} /2
\end{equation*}
Hierarchy is thus parameterized by the steepness ($\color{Purple}{s}$), centering ($\color{blue}{c}$), and amplitude ($\color{ForestGreen}{\Delta \tau}$) factors. 
The resulting shapes of the time constant hierarchy are shown in Figure~\ref{fig:Figure2}e. 
The effect of the steepness and centering parameters are analyzed while fixing the hierarchy amplitude at ${\color{ForestGreen}{\Delta \tau}}= \SI{150}{\milli \second}$ in Figure \ref{fig:Figure2}e, where 5-hidden layers feed-forward \acp{SNN} are trained on the SHD task. The average time constant $\tau_{\mu}$ is set to \SI{200}{\milli\second} as it maximizes the performance (see Supplementary Figure~\ref{fig:FigureS1}b). 
We show in a color map (Figure \ref{fig:Figure2}f) the classification accuracy with temporal hierarchy, compared to the case ${\color{ForestGreen}{\Delta \tau}}=\SI{0}{\second}$. 
Notably, all values are positive, confirming that temporal hierarchy correlates with a boost in performance. 
Certain combinations of steepness and centering parameters provide a larger increase in classification accuracy, with a peak of 4.1\% when ${\color{Purple}{s}}=0.5$ and ${\color{blue}{c}}=0.5$. 
The tanh shapes that yield the best performance are illustrated in Supplementary Figure \ref{fig:FigureS1}d. Fixing this combination of optimal hierarchy parameters, the number of neurons per hidden layer is increased to 128 neurons, and the results are compared with and without temporal hierarchy for increasingly deeper networks (Figure~\ref{fig:Figure2}f). 
Temporal hierarchy improves performance for both shallow and deep \acp{SNN}. Interestingly, the \acp{SNN} with 32 neurons per hidden layer with temporal hierarchy perform on par with larger \acp{SNN} with 128 neurons per hidden layer and homogeneous time constants. 
This is a considerable achievement, as the latter network has 6 times more parameters. From a hardware-centered perspective, temporal hierarchy allows shrinking the size of an SNN model while granting the same level of performance. 
For an estimation of the effect of the hierarchy amplitude $\color{ForestGreen}{\Delta \tau}$ on the performance, refer to Supplementary Figure~\ref{fig:FigureS1}c.\\
To sum up, we report the results on the \ac{SHD} dataset with the temporal hierarchy-endowed \acp{SNN} (h\acp{SNN}) in Table \ref{table1}. 
Furthermore, the hierarchy of time constants is tested on the Spiking Speech Command (SSC) dataset \cite{warden2018speech}, showing a performance boost compared with a network with homogeneous time constants. These results are found in Table \ref{table2}.

\subsection{Optimization discovers time constant hierarchy} \label{optimize_tau}

\begin{figure}[t!]
    \centering
    \includegraphics[width=0.99\linewidth]{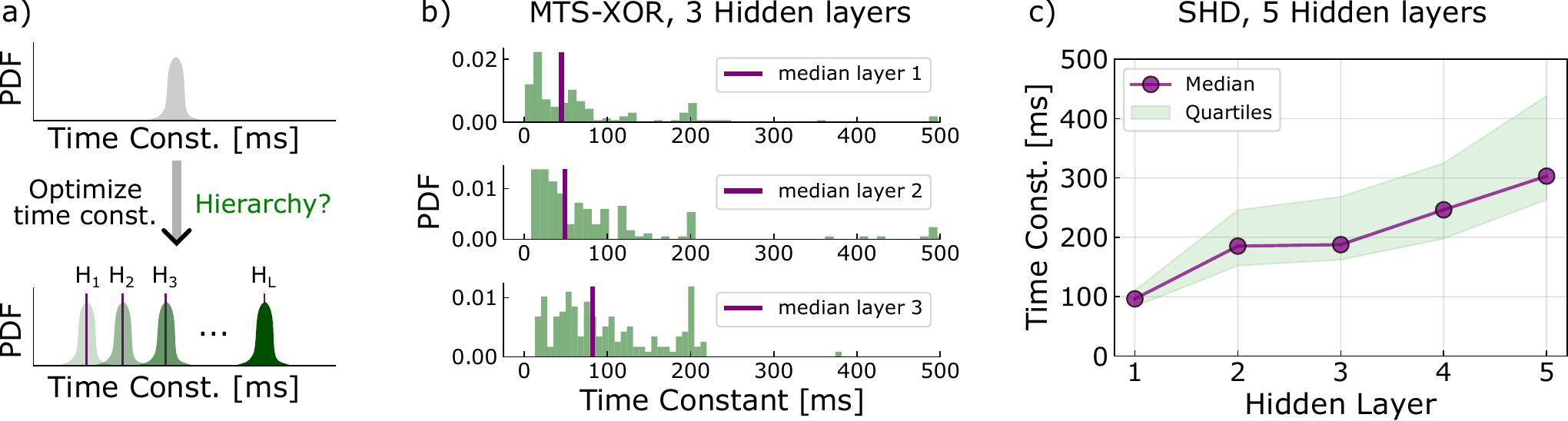}
    \caption{Optimizing the time constants in \acp{SNN}. a) Multi-layer \acp{SNN} are initialized with the same distribution of time constants through the hidden layers. Each time constant is optimized individually. b) Probability Density Function for the optimized time constants in a 3-hidden-layer SNN solving the MTS-XOR task. The median time constant grows in each layer, indicating the formation of a hierarchy. c) Mean time constant from 5-hidden-layer \acp{SNN} solving the SHD task. The mean grows indicating the hierarchy through the network. Results are averaged from 5 trials in both b) and c).}
    \label{fig:Figure3}
\end{figure}

The experiments in the previous section prove the efficacy of temporal hierarchy in the initialization of time constants of \ac{LIF} neurons as an inductive bias to \acp{SNN}. 
However, time constants were not optimized in these experiments. Here we observe the results of optimizing the time constants of \ac{LIF} neurons in \acp{SNN}, looking for a pattern in these trained values. %
To find out whether a hierarchy of time constants naturally emerges from optimization, we initialize \acp{SNN} with a neuron time constant drawn from a Gaussian distribution with a given mean ($\tau_{\mu}$) and standard deviation of 20\% of the mean. 
Such initialization is the same for all the hidden layers in the \acp{SNN}. 
Following a conventional gradient descent approach, we update the time constant of hidden neurons while training the \acp{SNN} to solve the MTS-XOR and \ac{SHD} tasks (Fig~\ref{fig:Figure3}a). 
For more details on the optimization of time constants, refer to Method section \ref{loss_regularization}.
The probability density function (PDF) of the time constant in the hidden layer of 5 \acp{SNN} trained on the MTS-XOR is shown in Figure~\ref{fig:Figure3}b. 
As previously observed in \cite{perez2021neural}, the distribution of optimized time constants (PDFs in green) in \acp{SNN} follows fat-tailed log-normal distributions. 
Along with the PDF, we indicate the median of the distributions in purple. Notably, the median values grow from layer to layer in the \ac{SNN}. 
While the probability density function (PDF) appears similar in the first and second hidden layers, the PDF in the third layer features a larger tail, causing the median value to be larger. 
This confirms that the hierarchy of time constants is generated from optimization.\\
As a further confirmation, we repeated the experiment with 5-hidden layer \acp{SNN} trained on the SHD task. Again, the methodology follows what is described in Fig~\ref{fig:Figure3}a, and we show the resulting mean values of the time constant distribution per hidden layer. Once again, the mean value grows through the layers indicating that the optimization results in a temporal hierarchy formed by the time constants. 

\subsection{Temporal hierarchy in convolutional SNN} \label{hierarchy_conv}

\begin{figure}[t!]
    \centering
    \includegraphics[width=0.99\linewidth]{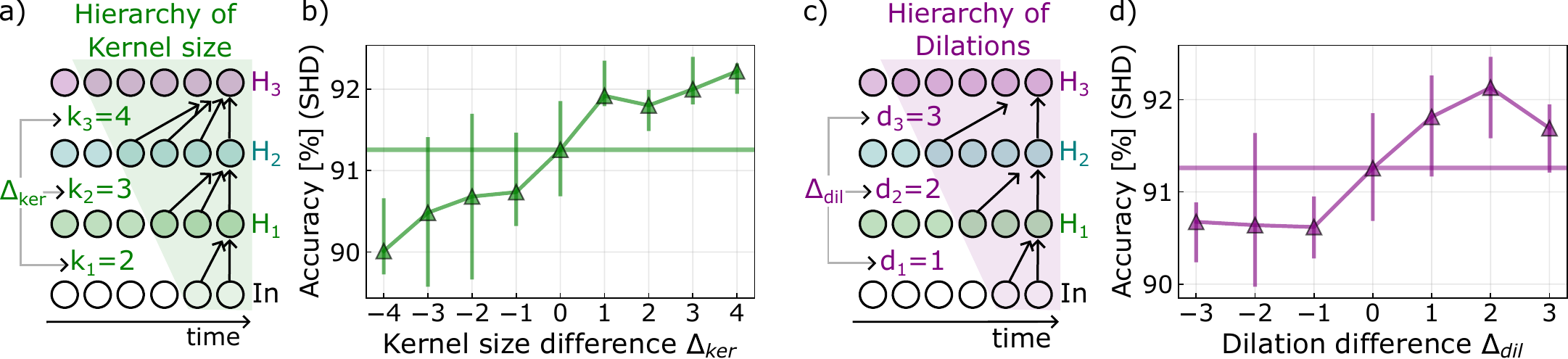}
    \caption{Temporal hierarchy in convolutional Spiking Neural Networks. a) Kernel size can vary in 1D-Causal convolution. One can increase the kernel size across hidden layers, forming a hierarchy of temporal representation. b) Effect of the hierarchy of kernel size on the SHD task. Positive hierarchy increases the classification accuracy. c) Dilation in Causal Temporal Convolution controls the size of the receptive field. One can construct a temporal hierarchy by progressively enlarging the dilation across the hidden layers. d) Effect of the hierarchy of dilations on the SHD task. Incrementing dilation through the hidden layers improves the performance of the network. Error bars show the quartiles over 10 trials.}
    \label{fig:Figure4}
\end{figure}

Delays and Temporal Causal Convolutions have been demonstrated extremely effective in improving the expressiveness of \acp{SNN} \cite{d2024denram,hammouamri2023learning,sun2023adaptive,sun2022axonal, goltz2024delgrad, deckers2024co}. 
In particular, DenRAM \cite{d2024denram} introduces synaptic delays in dendritic-like branches that can be modeled as temporal convolutions with a kernel size of 1 and variable - though fixed after initialization - dilation, showing superior performance with minimized hardware resources and parameter count. 
These results are extended in \cite{hammouamri2023learning} where multi-layer \acp{SNN} are equipped with temporal causal convolutions whose dilation can be learned. 
These works either assume shallow \acp{SNN} \cite{d2024denram} or do not differentiate between the convolutions in different layers \cite{hammouamri2023learning} with specific initialization or training policies. 
Here, we perform experiments with kernel size and dilation in temporal convolutions in \acp{SNN}, demonstrating that imposing temporal hierarchy can yield better performance. 
Preliminary results on kernel size and dilation hyper-parameter optimization are shown in Supplementary Figure \ref{fig:FigureS3}.\\
The kernel size in a temporal convolution corresponds to the size of the convolutional window, i.e. the number of delay parameters used to compute a given neuron's activation, as illustrated in Figure~\ref{fig:Figure4}a. 
We construct a temporal hierarchy by increasing the kernel size at every layer, denoting as $\Delta_\mathrm{ker}$, the kernel size difference across the network, i.e. the amplitude of the temporal hierarchy. 
Based on a network of two hidden layers, with 128 neurons per hidden layer, dilation fixed at five, trained to solve the \ac{SHD} task, we analyze the effect of temporal hierarchy constructed with the kernel size in Figure~\ref{fig:Figure4}b. 
While the mean kernel size is five, we sweep $\Delta_\mathrm{ker}$ between $-4$ and $+4$, noting that performance positively correlates with the amplitude of temporal hierarchy. This confirms the trend observed through the experiments in Fig.~\ref{fig:Figure2}.\\
The dilation in temporal convolution determines the size of the receptive fields of post-synaptic neurons, i.e. the extent of the delays to compute the neuron's activation, as illustrated in Figure~\ref{fig:Figure4}c. 
Temporal hierarchy is here constructed by progressively increasing dilation through the layers of the SNN. 
We construct a temporal hierarchy by increasing the dilation at every layer, denoting the dilation size difference across the network as $\Delta_\mathrm{dil}$. The effect of such parameters on classification performance on the \ac{SHD} datasets is shown in %
Figure~\ref{fig:Figure4}d demonstrates a positive correlation between classification accuracy and temporal hierarchy. This suggests once again that incorporating temporal hierarchy may serve as an effective inductive bias for improving \ac{SNN} performance.
The kernel size and dilation hierarchy are then combined, along with data augmentation, to maximize the performance of \acp{SNN}. 
The results are summarized in Table \ref{table1}, where we show how temporal hierarchy allows \acp{SNN} to reach competitive performance compared with the state-of-the-art. 
The best performing convolutional \ac{SNN} features $\Delta_\mathrm{ker}=3$ and $\Delta_\mathrm{dil}=2$.\\
Note that other factors play a crucial role in significantly enhancing performance. The type of data augmentation used in the experiments is detailed in the Methods section. Specifically, the ``Test as Valid'' column in Table \ref{table1} merits further discussion.
The \ac{SHD} dataset is originally split into a training and test set, without a validation set. 
Common practice is to split the training set into two sub-sets, with a conventional 80\%-20\% split, forming a validation set. 
This method is utilized in \cite{d2024denram, nowotny2022loss} and we also recommend this approach. 
However, multiple publications feature another validation strategy, where the test set is assumed as the validation set, and performance (along with hyper-parameter optimization and validation early-stopping) is evaluated on the test set. As this method leaks information from the test set into the training phase, the generalization capabilities of the model can no longer be evaluated properly. With the sole scope of comparing performance with works utilizing such validation scheme \cite{hammouamri2023learning, schone2024scalable}, we also report results obtained with this method (``Test as Valid''). A thorough comparison with the state-of-the-art in SHD is reported in the Supplementary Information (Table~\ref{tableS1}).\\
Additionally, we performed experiments on the SSC dataset with temporal convolution hierarchy, once again observing a slight increase in performance. Results are reported in Table \ref{table2}.

\bgroup
\def\arraystretch{1.2}
\begin{table}[]
\centering
\resizebox{0.70\textwidth}{!}{%
\begin{tabular}{lccccccc}
 \hline & Hierarchy & \begin{tabular}[c]{@{}c@{}}Data \\ Augment\end{tabular} & \begin{tabular}[c]{@{}c@{}}Test as \\ Valid*\end{tabular} & \begin{tabular}[c]{@{}c@{}}\# hid.\\ layers\end{tabular} & \begin{tabular}[c]{@{}c@{}}hidden\\ size\end{tabular} & \# params & Accuracy\% \\ \hline
\multirow{4}{*}{hSNN} &  &  &  & 3 & 32 & 25k & 81.2 ± 1.4 \\
 & \checkmark &  &  & 3 & 32 & 25k & 85.6 ± 1.2 \\
 &  &  &  & 4 & 128 & 150k & 86.4 ± 1.3 \\ %
 & \checkmark &  &  & 4 & 128 & 150k & 88.1 ± 0.6 \\ \hline
\multirow{4}{*}{hCSNN} &  &  &  & \multirow{4}{*}{4} & \multirow{4}{*}{128} & \multirow{4}{*}{345k} & 91.2 ± 0.8 \\
 & \checkmark &  &  &  &  &  & 92.2 ± 1.1 \\
 & \checkmark & \checkmark &  &  &  &  & 94.1 ± 0.7 \\
 & \checkmark & \checkmark & \checkmark &  &  &  & 94.7 ± 0.7 \\ \hline
\end{tabular}%
}
\vspace{0.2cm}
\caption{Performance with and without temporal hierarchy of \acp{SNN} on the Spiking Heidelberg Digits dataset.}
\label{table1}
\end{table}

\bgroup
\def\arraystretch{1.2}
\begin{table}[]
\centering
\resizebox{0.65\textwidth}{!}{%
\begin{tabular}{lcccccc}
\hline
 & Hierarchy & \begin{tabular}[c]{@{}c@{}}Data \\ Augment\end{tabular} & \begin{tabular}[c]{@{}c@{}}\# hid.\\ layers\end{tabular} & \begin{tabular}[c]{@{}c@{}}hidden\\ size\end{tabular} & \# params & Accuracy\% \\ \hline
\multirow{2}{*}{hSNN} &  &  & \multirow{2}{*}{2} & \multirow{2}{*}{128} & \multirow{2}{*}{150k} & 71.0 ± 1 \\
 &\checkmark &  &  &  &  & 73.1 ± 0.3 \\
\multirow{2}{*}{hCSNN} &  &\checkmark & \multirow{2}{*}{2} & \multirow{2}{*}{256} & \multirow{2}{*}{1M} & 78.2 ± 0.1 \\
 &\checkmark &\checkmark &  &  &  & 79.2 ± 0.2 \\ \hline
\end{tabular}%
}
\vspace{0.2cm}
\caption{Performance with and without temporal hierarchy of \acp{SNN} on the Spiking Speech Command dataset.}
\label{table2}
\end{table}

\section{Discussion} \label{sec:discussion}
This work proposes to organize temporal computational elements in a hierarchy, from fast to slow, through the layers of \acp{SNN}. Temporal hierarchy is constructed with two different computational elements: neuronal dynamics, i.e.\ time constants in \ac{LIF} neurons, and delays, i.e.\ temporal causal convolutions. 
First, we show that organizing time constants in a hierarchy yields a performance increment in common temporal benchmark tasks. Then we let optimization learn the time constants in \acp{SNN}, finding that hierarchy emerges from the process. The benefits of temporal hierarchy are also found in temporal causal convolutional \acp{SNN}, where hierarchy is constructed from the kernel size and dilation of the convolution. In all these cases, we benchmark \acp{SNN} on temporal tasks, where the evolution of the data over time encodes relevant information.

As a counter-example, we investigate the case where the input data is static, i.e. it does not change over time, as in the case of images. In Supplementary Figure \ref{fig:FigureS4} we classify MNIST digits with \acp{SNN}, showing that temporal hierarchy plays no role in this case. This sets a boundary to the practical utility of temporal hierarchy, which is beneficial only if the data features relevant temporal information. As further proof, we change the encoding of MNIST images, this time in the common ``sequential MNIST'' (sMNIST \cite{le2015simple}) setting: the information from each pixel is passed to an SNN sequential. In this setting, the sequence encodes relevant information, and temporal hierarchy becomes beneficial again, providing a boost of classification accuracy.

\paragraph*{Performance benefit} %
In the experiments on temporal hierarchy constructed with time constants (Table \ref{table1}), the benefit in the performance of the \ac{SHD} task was  4.1\% with a small size \ac{SNN} (32 neurons per hidden layer) and it reduced to 1.5\% with a larger \ac{SNN} (128 neurons per hidden layer). When constructing temporal hierarchy with convolutions the benefit of hierarchy on the \ac{SHD} task further decreased to 1\% (Table \ref{table1}). We also show how data augmentation can yield a comparable performance boost, raising the question of the true power of temporal hierarchy. While we acknowledge that the performance benefit is not drastic, the power of temporal hierarchy is in its simplicity and generality. It is simple to implement, as it acts as a rule for initialization. It is general as it has been proven beneficial for the most common computational primitives in \acp{SNN}, neuronal dynamics and delays. Furthermore, the experiments run in this work involve datasets with limited complexity. Possibly, testing on datasets with more complex temporal features could reveal even greater benefits in terms of performance.

\paragraph*{Implications for neuromorphic computing}
The hardware implementation of \acp{SNN} is the main aim of neuromorphic engineering. Many existing neuromorphic chips \cite{pehle2022brainscales, richter2024dynap, orchard2021efficient} support the \ac{LIF} neuron model, and some circuits to implement dendritic delays have been developed \cite{d2024denram, ramakrishnan_etal2013_suma, wang2010multilayer}. %
This makes temporal hierarchy a valuable feature that most neuromorphic hardware designers can easily leverage. 
For example, in Figure \ref{fig:Figure2}g, we show that temporal hierarchy allows reducing the size of an \ac{SNN} by $6 \times$, while maintaining the same performance on the \ac{SHD} task. In this manner, the energy consumption and memory footprint of a neuromorphic chip running an \ac{SNN} can be considerably reduced. 
Furthermore, temporal hierarchy has been demonstrated on multiple temporal processing primitives. This indicates that temporal hierarchy can be customized to the specifics of each neuromorphic chip implementation, providing a performance benefit independent of the chip's design choices.

\paragraph*{Outlook}
In addition to our demonstration of the benefit of temporal hierarchy in multiple settings, there are numerous ways to extend this work. On the one hand, more complex neuronal dynamics can be leveraged to form temporal hierarchies. The true potential of the dynamics of, for example, Izhikevich neurons \cite{izhikevich2001resonate} or multi-compartment models could be unlocked by a hierarchical setting of their hyper-parameters. For example, one could build a temporal hierarchy exploiting complex computational primitives of biologically plausible neuron models, such as plateau potentials, spike bursts, and intrinsic plasticity. On the other hand, the benefit of temporal hierarchy could be even greater when tested on datasets with a higher degree of temporal information. In some cases, temporal hierarchy is a natural feature of data. In speech, information is represented across multiple levels: phonemes, words, phrases, and semantics. A single model that tackles all these input representations will likely benefit from a temporal hierarchy.

\section{Methods}

\subsection{SNN implementation}
For all our experiments, we implement \acp{SNN} using the JAX \cite{jax2018github} python library to benefit from auto-differentiation and speed of execution. We run the experiments on two A6000 GPUs.
Spiking Neural Networks are formalized based on the Leaky-Integrate-and-Fire equation:
\begin{align*}
    \tau \dot{u}(t) & = -u(t) + f( I(t), W) - u_\mathrm{th} s(t) \\
    s(t) & = \theta ( u(t) - u_\mathrm{th} )
\end{align*}
where, $u(t) \in \mathcal{R}^{N}$ is the membrane voltage of the neuron, $\tau \in \mathcal{R}^{N}$ its time constant, $W \in \mathcal{R}^{M,N}~or~\mathcal{R}^{K,M,N}$ the input weights and $I \in \mathcal{R}^{M}$ the layer's inputs. $N$ is number of neurons, $M$ number of input connections, $K$ kernel size (when in convolution mode). The neuron emits a spike ($s(t) \in \mathcal{R}^{N}$) whenever the membrane voltage $u(t)$ crosses the threshold $u_\mathrm{th}$. We set $u_\mathrm{th}=1$ for all our experiments. After a spike, the membrane voltage is reset: we chose a ``soft reset'' mechanism, i.e. the membrane voltage is not immediately reset to zero, but it is decreased proportionally to the threshold voltage.\\
As it is well known, the non-linearity of a \ac{LIF} neuron (Heaviside function) is non-differentiable. To allow the gradient to flow, we use the Surrogate Gradient technique \cite{neftci2019surrogate}, replacing the derivative of the Heaviside function with a ``box function'', as in \cite{bittar2022surrogate}.\\
Based on the equations above, we build layers of spiking neurons connected by either fully connected layers 
\begin{equation*}    
    f(I(t), W) = I(t)\,W \quad \text{with} \quad W \in \mathcal{R}^{M,N}
\end{equation*}
or by 1D temporal causal convolution 
\begin{equation*}
    f(I(t), W) = I(t) * W \quad \text{with} \quad W \in \mathcal{R}^{K,M,N}
\end{equation*}
In the latter case, we pad the input sequence $I(t)$ with as many zeros as $K-1$ to make the temporal convolution causal.\\
In each layer, neurons are initialized with a certain time constant. We draw these time constants from a Gaussian distribution with a given mean and standard deviation of 20\% of the mean. In this way, we introduce a slight degree of heterogeneity in the SNN, although we observe minimal no beneficial impact on the performance unlike \cite{perez2021neural}. This can be explained by the fact that we do not feature recurrent connections, and thus do not exploit the recurrent dynamics that is enriched by the heterogeneity of neurons. Note that temporal hierarchy is formed by controlling the mean of the time-constant distribution across the layers.\\
Multiple such layers are concatenated to form either a multi-layer spiking perceptron (experiments in Figure \ref{fig:Figure2}) or a convolutional SNN (Figure \ref{fig:Figure4}). The output layer, instead, is always made of Leaky-Integrators neurons, i.e. non-spiking neurons. The time constant for such output neurons is \SI{200}{\milli\second} for all experiments. The values of other relevant hyperparameters can be found in Supplementary Table \ref{table_hyper}.

\subsection{Loss function and regularization} \label{loss_regularization}
\paragraph*{Loss Function} Concerning the loss function, all experiments involve supervised classification tasks, for which Cross-Entropy is chosen. The output logits are computed as in \cite{bittar2022surrogate}, thus computing the softmax of the output neurons membrane voltage at every time step, integrating these values over the whole duration of the task, and applying the softmax once again on the obtained values. The resulting loss function averages the cross-entropy over the task duration:

\begin{equation*}
    \mathcal{L}_\text{sum} = - \frac{1}{T} \sum_{t}^{T} \log\left( \frac{ \exp\left( \mathrm{out}[t, y] \right) } { \sum_{n}^{N} \exp\left( \mathrm{out}[t, n] \right) } \right)
\end{equation*}
where $T$ is the task duration, $N$ number of output neurons, $\mathrm{out} \in \mathcal{R}^{T, N}$ the output neurons membrane voltage, $y$ the desired label.\\
In the case of the MTS-XOR task, we opted for a ``max-over time'' loss function, typical in many \acp{SNN} in the literature \cite{cramer2020heidelberg, neftci2019surrogate}. This loss function looks for the maximum value for the membrane voltage of the output neuron over a certain period. In the case of the MTS-XOR, this period corresponds to the presentation of the cues in Channel B. The "max-over time" loss function equation is the following:
\begin{equation*}
    \mathcal{L}_\text{max} = - \log \left( \frac{\exp( \mathrm{max}_T\left(\mathrm{out}[y]) \right)}{\sum^N_n \exp( \mathrm{max}_T\left( \mathrm{out}[n] ) \right) } \right)
\end{equation*}
where $\mathrm{max}_T$ indicates the selection of the maximum value of the membrane voltage over the selection period $T$.\\

\paragraph*{Regularization} Dropout with a probability of 10\% is applied in the experiments in Figure \ref{fig:Figure2} while the probability is increased at 40\% in Figure \ref{fig:Figure4}. L2-regularization with a magnitude of $10^{-4}$ is utilized for experiments with the convolutional \acp{SNN}. When training the time constants of neurons (Fig.\ref{fig:Figure3}), a regularization term is applied to prevent unreasonable values. It is reminded that the time constant has to be positive and that values that are too large might lead to numerical instability. The resulting regularization term on the time constant is a re-scaled L2-norm:
\begin{equation*}
    \mathcal{L}_{\tau, L2} = \sum_l^{L} L2\left( \tau(l) - \mathrm{mean}[\tau(l)] \right)
\end{equation*}
where $\tau(l)$ is the time constant at a given hidden layer, with L hidden layers in total. 

\subsection{Data Augmentation}
In some experiments on the convolutional SNN section (Table \ref{table1}) we utilize data augmentation to improve the generalization performance of the model. However, in the spirit of this work, we looked for a simple yet effective strategy to augment the SHD and SSC datasets. We don't upsample the dataset but simply transform its samples. In particular, the only transformation we apply is a transposition of the input channels, equivalent to a frequency shift of the speaker's voice. For each data point, we draw a number with a mean of 10 and a standard deviation of 5, representing the frequency shift of the data. We then pad the input sequence with as many channels as the drawn frequency shift and truncate the exceeding channels to recover the original dataset input size (700 channels). The frequency shift can be positive or negative with 0.5 probability. Each sample in a batch of data has an independent frequency shift.

\section{Acknowledgment and Funding}
We want to thank the EIS-Lab, especially Tristan Torchet for assisting with the code implementation.
The presented work has received funding from the Swiss National Science Foundation Starting Grant Project UNITE (TMSGI2-211461).

\bibliographystyle{unsrtnat} %
\bibliography{references}  %

\setcounter{equation}{0}
\setcounter{figure}{0}
\setcounter{table}{0}
\makeatletter
\renewcommand{\theequation}{S\arabic{equation}}
\renewcommand{\thefigure}{S\arabic{figure}}
\renewcommand{\thetable}{S\arabic{table}}

\section*{Supplementary Information}

\paragraph*{Table of Hyperparameters}
We list in Table~\ref{table_hyper} the hyperparameters used in the experiments. The acronyms and symbols refer to: learning rate (LR), feed-forward (FF), convolutional (conv), Max-over-time loss function ($\mathcal{L}_\mathrm{max}$), sum-of-softmax-output $\mathcal{L}_\mathrm{sum}$, output neuron time constant $\tau_\mathrm{out}$, mean hidden layer time constant $\tau_{\mu}$.

\begin{table}[h]
\centering
\resizebox{\textwidth}{!}{%
\begin{tabular}{ccccccccccc}
\hline
 &  & Batch Size & Epochs & Dropout & LR, decay {[}epochs start, decay{]} & Initialization & Loss & $\tau_{\mu}$ [s] & $\tau_\mathrm{out}$ [s] \\ \hline
MTS-XOR & FF & 512 & 60 & 0.1 & 0.01, linear {[}25, 0.5{]} & Xavier & $\mathcal{L}_\mathrm{max}$ & 0.3 & 0.2 \\
\multirow{2}{*}{SHD/SSC} & FF & 256 & 60 & 0.1 & 0.01, linear {[}25, 0.5{]} & Xavier & $\mathcal{L}_\mathrm{sum}$ & 0.2 & 0.2 \\
 & Conv & 256 & 100 & 0.4 & 0.01, linear {[}25, 0.5{]} & Xavier & $\mathcal{L}_\mathrm{sum}$ & 0.2 & 0.2 \\ \hline
\end{tabular}%
\label{table_hyper}
}
\vspace{0.1cm}
\caption{Table of the hyperparameters.}
\end{table}

\subsection*{Feed Forward SNN with temporal hierarchy} \label{Supplementary_FF_SNN}

\begin{figure}[h!]
    \centering
    \includegraphics[width=0.95\linewidth]{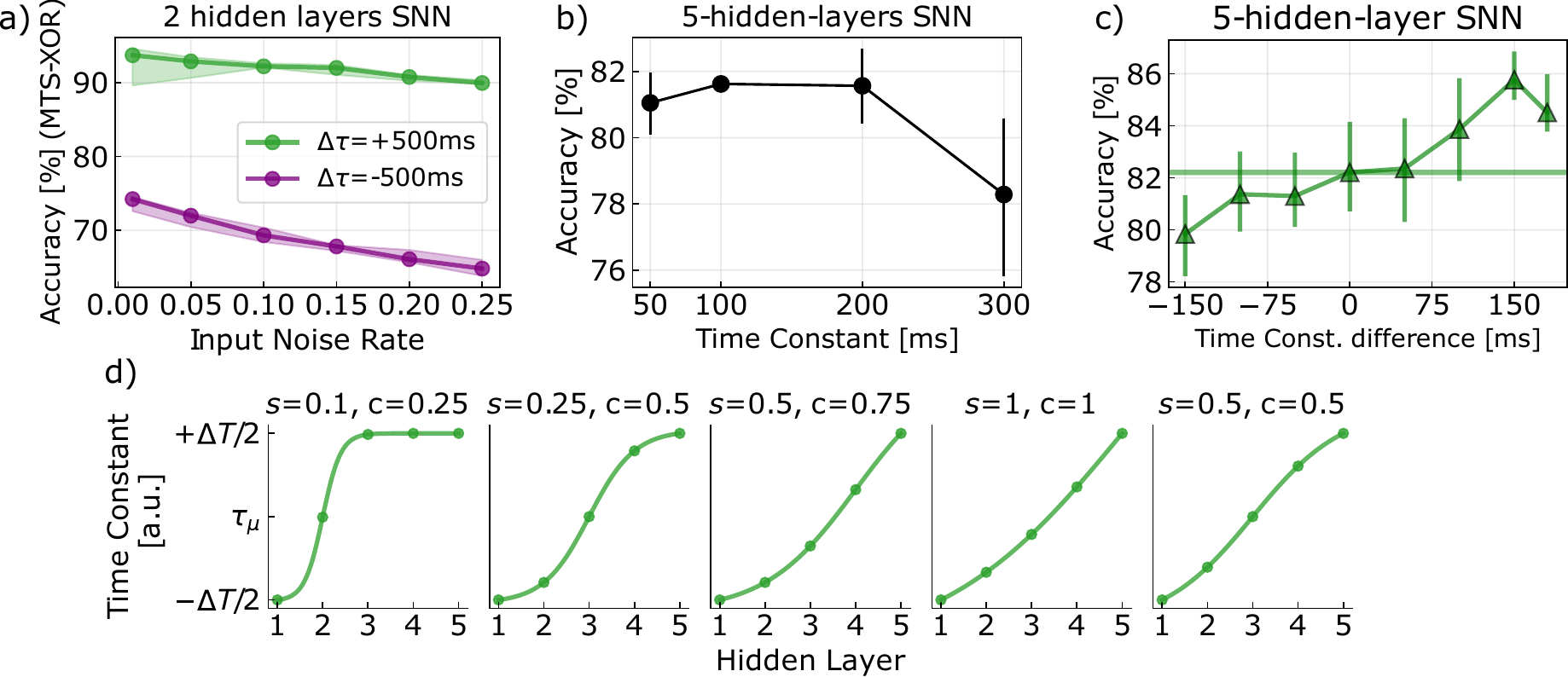}
    \caption{a) Accuracy of SNNs with positive ($\Delta \tau = \SI[retain-explicit-plus]{+500}{\milli\second}$) and negative ($\Delta \tau = \SI[retain-explicit-plus]{-500}{\milli\second}$) hierarchy of time constant, as a function of the background noise rate on the MTS-XOR task. Shaded areas show the quartiles over 5 trials. b) Test accuracy on the SHD task for a network with homogeneous time constants through the layers. c) Hierarchy shapes of the $\tanh$ function that maximize classification accuracy. Error bars show the quartiles over 5 trials in b) and c). d) The $\tanh$ function parameterized by the 5 combinations of steepness and centering that yield the best results in Figure~\ref{fig:Figure2}f.}
    \label{fig:FigureS1}
\end{figure}

The intuition behind the role of temporal hierarchy is that the first hidden layer acts as a filter, while deeper layers extrapolate higher-level features. To prove this, we set up a simple experiment trying to perform the MTS-XOR task with an increasing background noise rate. Figure \ref{fig:FigureS1}a shows that SNNs with positive hierarchy are more resilient to input background noise, while negative hierarchy leads to a slightly larger dip in classification accuracy.\\
To assess the performance of temporal hierarchy in feed-forward SNNs, we first optimize a conventional SNN by finding the optimal average time constant $\tau_{\mu}$, homogeneous through the layers in this case. Results of this characterization on the SHD dataset are shown in Supplementary Figure \ref{fig:FigureS1}b, where classification accuracy is maximized in the range $\tau_{\mu} \in [\SI{100}{\milli\second}, \SI{200}{\milli\second}]$. For the rest of the experiments, we thus assume $\tau_{\mu}=\SI{200}{\milli\second}$.\\
While in Figure~\ref{fig:Figure2}e we analyzed the effect of the shape of time constant hierarchy, in Supplementary Figure~\ref{fig:FigureS1}c we look at the effect of the amplitude $\color{ForestGreen}{\Delta \tau}$. As previously observed in Figure~\ref{fig:Figure2}d, there is a correlation between classification accuracy and temporal hierarchy amplitude, which can provide a boost of performance of up to 4.1\%.\\
Based on the analysis of the shape of hierarchy (Figure~\ref{fig:Figure2}e), we plot the combinations of steepness and centering that optimize the performance in Supplementary Figure~\ref{fig:FigureS1}c.\\

\subsection*{State-of-the-art SNNs in temporal tasks}

To better compare the performance of SNNs endowed with temporal hierarchy, we compile a table of results (Table \ref{tableS1}) on the Spiking Heidelberg Digit dataset \cite{cramer2020heidelberg}. This table is based on the resources in the SHD original website \url{https://zenkelab.org/resources/spiking-heidelberg-datasets-shd/}. The scope of the table is to highlight the role of the different computational principles and inductive biases on the performance of SNNs. Note that this table only includes spike-based algorithms. Other neural network architectures (LSTMs \cite{cramer2020heidelberg}, CNNs, and event-based State Space Models \cite{schone2024scalable}) have been tested on this task.\\
Hierarchy improves the performance of SNNs (hSNN), with the best result for hSNN less than 1\% away from the state-of-the-art performance \cite{hammouamri2023learning} despite featuring simple LIF-based convolutional SNNs. Note that in this work we wanted to isolate the effect of temporal hierarchy, thus the choice of basic SNN architectures, avoiding other features like recurrence \cite{cramer2020heidelberg}, adaptation \cite{bittar2022surrogate}, temporal attention, and complex neuron models.\\
We believe temporal hierarchy can easily be adopted by other researchers and improve the computational power of their models.

\begin{table}[h!]
\resizebox{\textwidth}{!}{%
\begin{tabular}{lccccccccccc}
 \hline & \multicolumn{5}{c}{Network Features} & \multirow{2}{*}{\begin{tabular}[c]{@{}c@{}}data \\ augment.\end{tabular}} & \multirow{2}{*}{\begin{tabular}[c]{@{}c@{}}Test as\\ Valid.\end{tabular}} & \multicolumn{3}{c}{Network Architecture} & Accuracy \\
 & Rec & Adapt & Att & \begin{tabular}[c]{@{}c@{}}delays/\\ conv.\end{tabular} & hierarchy &  &  & \begin{tabular}[c]{@{}c@{}}\# hid.\\ layers\end{tabular} & \begin{tabular}[c]{@{}c@{}}hidden\\ size\end{tabular} & \# params & \% \\ \hline
\begin{tabular}[c]{@{}l@{}}Perez-Nieves et al. \cite{perez2021neural} \end{tabular} & \checkmark &  &  &  &  &  &  & 1 & 256 & 100k & 82.7 ± 0.8 \\

\begin{tabular}[c]{@{}l@{}}D’Agostino et al. \cite{d2024denram} \end{tabular} &  &  &  & \checkmark &  &  &  & / & / & 224k & 90.7 \\
\begin{tabular}[c]{@{}l@{}}Yao et al. \cite{yao2021temporal} \end{tabular} &  &  & \checkmark &  &  &  &  & 2 & 128 & / & 91.1 \\
\begin{tabular}[c]{@{}l@{}}Sun et al. \cite{sun2023adaptive} \end{tabular} &  &  &  & \checkmark &  &  &  &  &  & 110k & 92.5 \\
\begin{tabular}[c]{@{}l@{}}Nowotny et al. \cite{nowotny2022loss} \end{tabular} & \checkmark &  &  &  &  & \checkmark &  & 3 & 256 & 250k & 93.5 ± 0.7 \\
\begin{tabular}[c]{@{}l@{}}Bittar and Garner \cite{bittar2022surrogate} \end{tabular} & \checkmark & \checkmark &  &  &  &  & \checkmark & 2 & 1024 & 1.8M & 94.6 \\
\begin{tabular}[c]{@{}l@{}}Hammouamri et al. \cite{hammouamri2023learning} \end{tabular} &  &  &  & \checkmark &  &  & \checkmark & 2 & 256 & 200k & 95.1 ± 0.3 \\ \hline
\multirow{2}{*}{\textbf{hSNN} {[}ours{]}} &  &  &  &  & \checkmark &  &  & 4 & 32 & 25k & 84.9 ± 1.6 \\
 &  &  &  &  & \checkmark &  &  & 4 & 128 & 150k & 88.1 ± 0.6 \\ \hline
\multirow{3}{*}{\textbf{hCSNN} {[}ours{]}} &  &  &  & \checkmark & \checkmark &  &  & \multirow{3}{*}{4} & \multirow{3}{*}{128} & \multirow{3}{*}{345k} & 92.2 ± 1.1 \\
 &  &  &  & \checkmark & \checkmark & \checkmark &  &  &  &  & 94.1 ± 0.7 \\
 &  &  &  & \checkmark & \checkmark & \checkmark & \checkmark &  &  & & 94.7 ± 0.7 \\ \hline
\end{tabular}%
}
\vspace{0.2cm}
\caption{Summary of the performance of SNNs with and without temporal hierarchy and comparison with the state-of-the-art in the SHD dataset. Abbreviations: Rec for recurrent, Adapt for neuronal adaptation, Att for temporal attention, conv. for temporal convolutions, ``Test as Valid.'' refers to utilizing the test set as the validation set for hyperparameter optimization, which is considered bad practice in machine learning but employed here once nevertheless to compare with other papers \cite{bittar2022surrogate, schone2024scalable}.}
\label{tableS1}
\end{table}

\subsection*{Optimization of the time constants discovers temporal hierarchy}
We extend here the results of the optimization of the time constant, in Figure \ref{fig:Figure3}. This time, we train a 5-hidden-layer SNN initialized with a homogeneous time constant of $\tau_{\mu}=\SI{200}{ms}$ on the Spiking Heidelberg Digit dataset. The resulting probability density functions averaged over 5 trials, are shown in Supplementary Figure \ref{fig:FigureS2}. Optimization produced nicely separated PDFs for the time constants of the neurons in each hidden layer.

\begin{figure}[h!]
    \centering
    \includegraphics[width=0.38\linewidth]{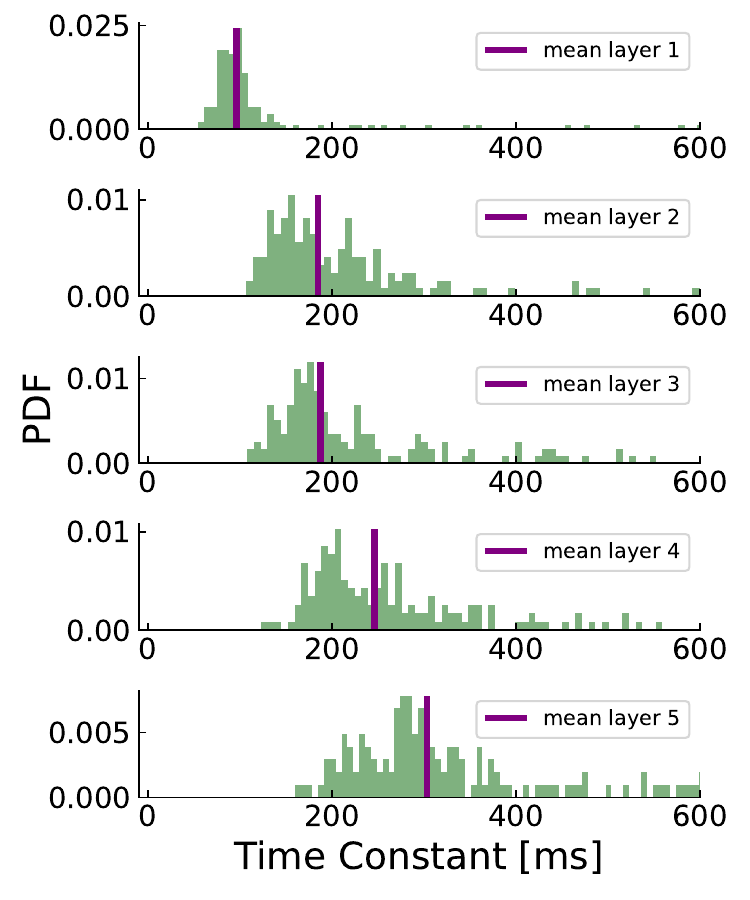}
    \caption{Optimization of the time constant training SNNs on the SHD task. The probability density function (PDF) of the time constants of neurons is shown in each of the 5 hidden layers (green). The mean value in each layer (purple) reveals the formation of hierarchy in the time constant. Results were collected by averaging from 5 trials.}
    \label{fig:FigureS2}
\end{figure}

\subsection*{SNNs with temporal causal convolutions}
As a preliminary analysis of temporal causal convolutional SNNs, the optimal kernel size and dilation are found via hyper-parameter optimization. In this experiment, data augmentation is disabled. In Figure \ref{fig:FigureS3} (left), we fix the dilation to 5 and vary the kernel size between 3 and 9. A kernel size of 5 optimizes the performance of the convolutional SNN. Similarly, Figure \ref{fig:FigureS3} (right) shows the classification accuracy as a function of the dilation, when the kernel size is fixed to 5. Despite a dilation of 3 optimizes performance, we fix the dilation to 5 for the rest of the experiments in Section \ref{hierarchy_conv}. This allows more control of the temporal hierarchy produced with dilations through the hidden layers.

\begin{figure}[h!]
    \centering
    \includegraphics[width=0.65\linewidth]{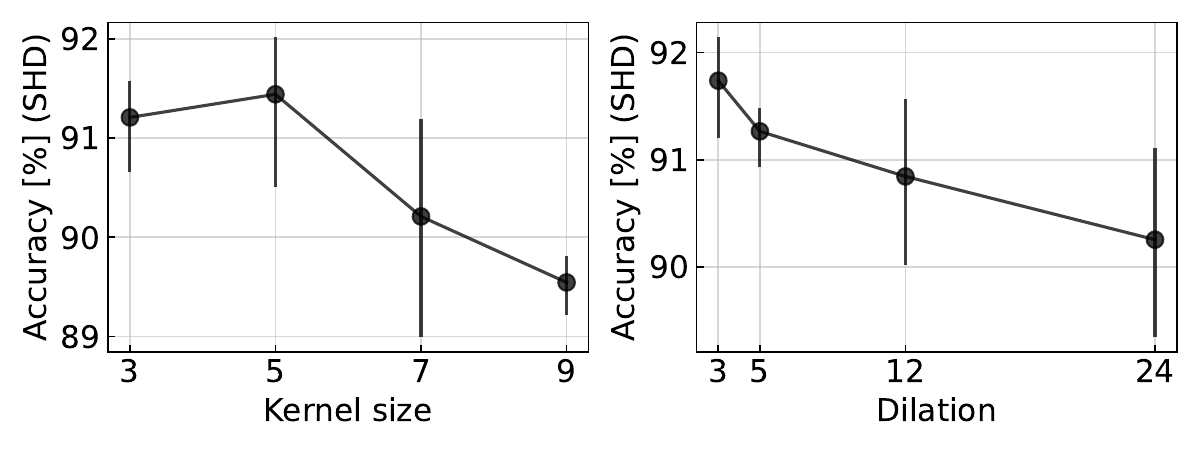}
    \caption{Hyperparameter tuning of the convolutional SNNs. We experiment on the effect of kernel size on the left, of the dilation on the right. Error bars show the quartiles over 5 trials.}
    \label{fig:FigureS3}
\end{figure}

\subsection*{Temporal hierarchy with static data}
While temporal hierarchy is demonstrated as a positive Inductive Bias that improves the performance in SNNs with tasks that feature information in the temporal domain, we investigate here what happens when the data is static in time, i.e. there is no relevant temporal dynamics at the input. To analyze this, temporal hierarchy is tested with the ever-green MNIST dataset \cite{lecun1998mnist}, a simple image classification dataset. As the data is provided as static images, we add a fictitious temporal dimension via Latency Coding (Figure~\ref{fig:FigureS4}a). This means the images are first flattened to 1 dimension (of size 784). Then the brightness of each pixel is converted to the latency of an event (or spike) that occurs at the input. Each image, originally in gray-scale with $28 \times 28$ size, is transformed to size $784\times T$, where $T$ is the number of time steps for the Latency Coding. For our experiments, we set $T=50$ steps, with each step $dt=\SI{20}{\milli\second}$, thus showing each image for 1 second. Note that summing along the fictitious temporal dimension would allow us to recover the information from the original image, provided that the precision of the input pixel brightness (8b) matches the precision of the temporal dimension ($T=2^{8}=256$). In this case, then, the input images are slightly compressed.\\
We first analyze the effect of setting the time constant homogeneously through the 2 hidden layers of a Spiking Neural Network. In Figure~\ref{fig:FigureS4}b we observe that, according to the expectation, the task can be solved by ``integrating'' the information at the input. Thus, the larger the time constant the better.\\
Setting $\tau_{\mu}=\SI{500}{\milli\second}$, we then impose a hierarchy of time constant with amplitude $\Delta \tau$ (Fig.~\ref{fig:FigureS4}c). Once again, changing the time constants does not impact the classification accuracy. This confirms once again that it is sufficient to integrate the information over time to solve this task. If the time constant of the neuron is larger than some threshold value (around \SI{200}{\milli\second} in this case), imposing a structure to the time constant does not affect the computation of the SNN. Temporal hierarchy is then a positive inductive bias for temporal tasks only.

\begin{figure}[h!]
    \centering
    \includegraphics[width=0.95\linewidth]{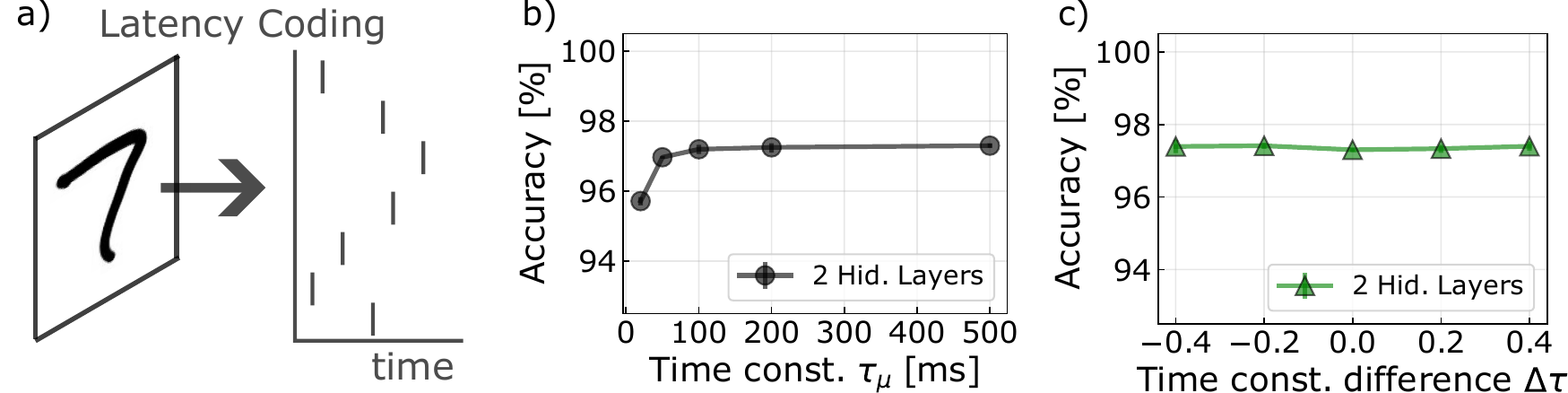}
    \caption{Role of temporal hierarchy as time constant hierarchy on static datasets. a) MNIST is converted with Latency Coding to introduce the temporal dimension to the data. b) Classification accuracy as a function of the mean homogeneous time constant $\tau_{\mu}$, same for all layers. c) Classification accuracy when time constant hierarchy is introduced, as a function of the hierarchy amplitude $\Delta \tau$. Error bars show the quartiles over 5 trials.}
    \label{fig:FigureS4}
\end{figure}

It is well-known that MNIST can be transformed into a sequential task, where the order of the sequence encodes information \cite{le2015simple}. Presenting the MNIST images as a 1-dimensional sequence (Figure \ref{fig:FigureS5}) offers a challenge for a classifying model, that has to encode the sequence into working memory to be able to classify it correctly. In this setting, temporal hierarchy should play a role, as it improves the temporal processing performance of SNNs. We test the hypothesis with an SNN composed of 2 layers with 64 LIF neurons each. Time is fictitious in this task, and we assign the sequence length (784 steps) to a duration of \SI{1}{\second}. The mean time constant $\tau_{\mu}$ of neurons is set 50 time-steps, corresponding to about \SI{60}{\milli\second}. Based on this network configuration, temporal hierarchy is introduced as the time constant difference $\Delta \tau$ in the initialization. The results of varying $\Delta \tau$ between \SI[retain-explicit-plus]{-100}{\milli\second} and \SI[retain-explicit-plus]{+100}{\milli\second} are shown in Figure \ref{fig:FigureS5}b: we note the positive correlation between $\Delta \tau$ and classification accuracy. However, accuracy improved also $\Delta \tau = \SI[retain-explicit-plus]{-100}{\milli\second}$. A possible explanation is that some longer time scales are still beneficial in solving this task, as evidenced in \cite{bellec2018long}. Similarly, we repeated the same experiment but with enabled recurrent connections (Figure \ref{fig:FigureS5}c). Classification accuracy had a positive impact, while the overall behavior related to the time-constant hierarchy remained.

\begin{figure}[h!]
    \centering
    \includegraphics[width=0.95\linewidth]{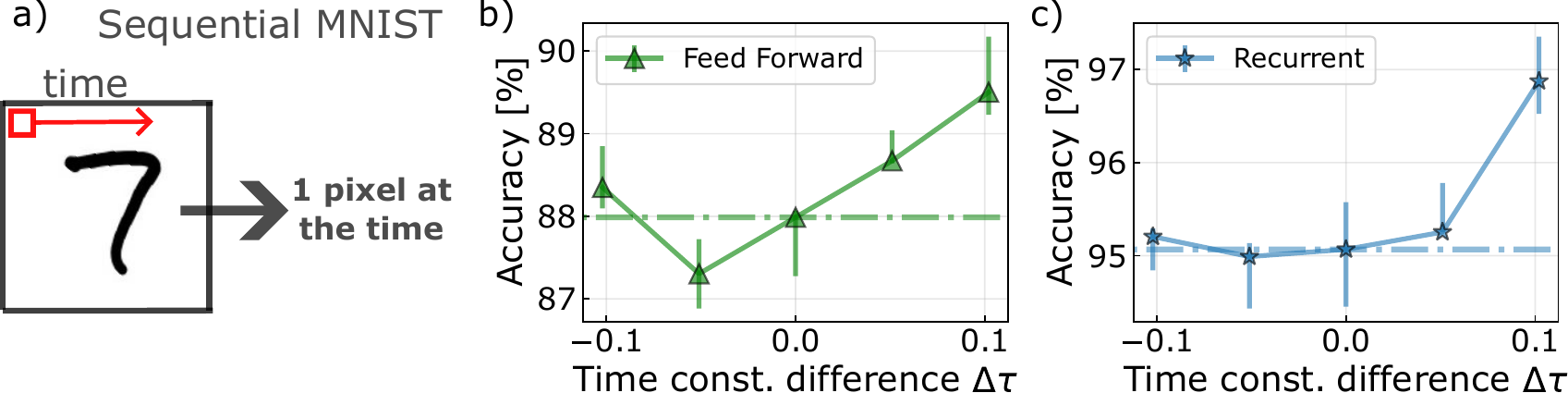}
    \caption{Role of temporal hierarchy as time constant hierarchy on the sequential MNIST task. a) MNIST is converted to a 1-dimensional sequence, with time stamps of 1~ms each. b) Classification accuracy when time constant hierarchy is introduced, as a function of the hierarchy amplitude $\Delta \tau$, in a feed-forward SNN. c) Same as b), but with enabled recurrent connections. Error bars show the quartiles over 5 trials.}
    \label{fig:FigureS5}
\end{figure}

\end{document}